\RequirePackage{amsmath}
\documentclass[runningheads]{llncs}

\usepackage[T1]{fontenc}
\usepackage{graphicx}
\usepackage{amssymb}
\usepackage[most]{tcolorbox}
\usepackage{lipsum}
\usepackage{xcolor}

\usepackage[colorlinks=true,linkcolor=blue,citecolor=blue,urlcolor=magenta]{hyperref}

\definecolor{babypink}{rgb}{1,0.9333,0.9529}
\definecolor{pink}{rgb}{1,0.7333,0.86}
\definecolor{pink2}{rgb}{1,0,1}
\usepackage{tikz,xcolor,hyperref}

\definecolor{lime}{HTML}{A6CE39}
\DeclareRobustCommand{\orcidicon}{
	\begin{tikzpicture}
	\draw[lime, fill=lime] (0,0) 
	circle [radius=0.16] 
	node[white] {{\fontfamily{qag}\selectfont \tiny ID}};
	\draw[white, fill=white] (-0.0625,0.095) 
	circle [radius=0.007];
	\end{tikzpicture}
	\hspace{-2mm}
}
\foreach \x in {A, ..., Z}{\expandafter\xdef\csname orcid\x\endcsname{\noexpand\href{https://orcid.org/\csname orcidauthor\x\endcsname}
			{\noexpand\orcidicon}}
}
			
\usepackage{tikz}
\usepackage{xparse}   

\DeclareRobustCommand{\authorpic}[2][5mm]{%
  \tikz[baseline={([yshift=-.25ex]current bounding box.center)}]{%
    \clip (0,0) circle (#1);
    \pgfmathsetlengthmacro{\picside}{sqrt(2)*#1}%
    \node at (0,0) {\includegraphics[width=\picside,height=\picside,keepaspectratio]{#2}};
    \draw[line width=0.4pt, color=white] (0,0) circle (#1);
  }%
}

\NewDocumentCommand{\AuthorWithPic}{O{5.5mm} O{0.20em} m m}{%
  \texorpdfstring{\authorpic[#1]{#4}\kern #2}{}%
  #3%
}

\usepackage{xcolor}
\definecolor{linkpinkix}{HTML}{EA335A} 

\definecolor{linkpink}{HTML}{EA335A}

\newcommand{\shadedlink}[2]{%
  \tikz[baseline=(n.base)]\node[
    fill=linkpink,
    fill opacity=0.5,
    text opacity=1,
    rounded corners=.3ex,
    inner xsep=.35em,
    inner ysep=.15em
  ] (n) {\href{#1}{\textcolor{blue!70!black}{#2}}};%
}

\begin{document}
\nocite{*}

\title{Multi-Sensory Cognitive Computing for Learning Population-level Brain Connectivity}
\author{
\AuthorWithPic[6mm][0.18em]{Mayssa Soussia}{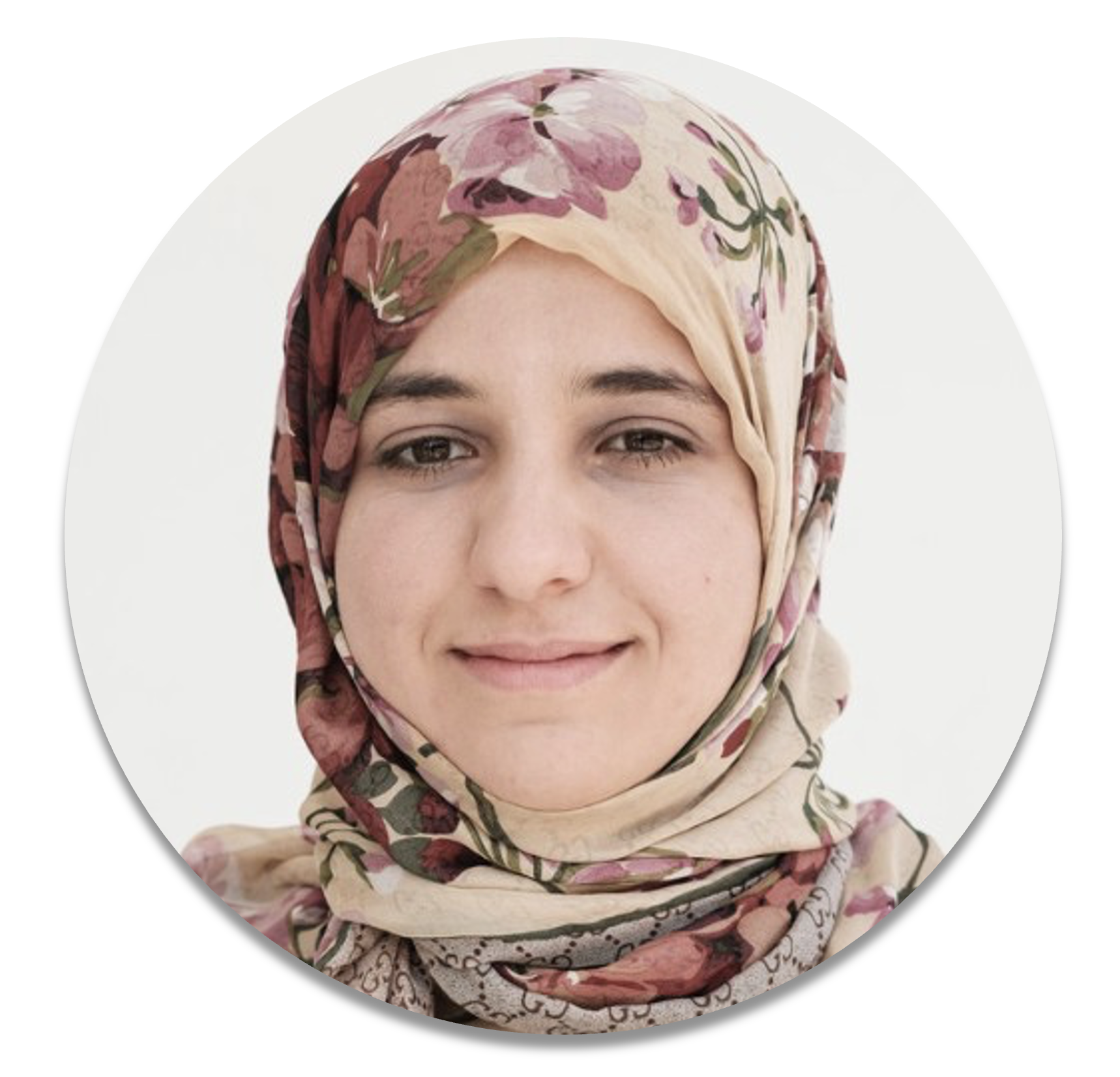}\inst{1,2} \and
\AuthorWithPic[6mm][0.18em]{Mohamed Ali Mahjoub}{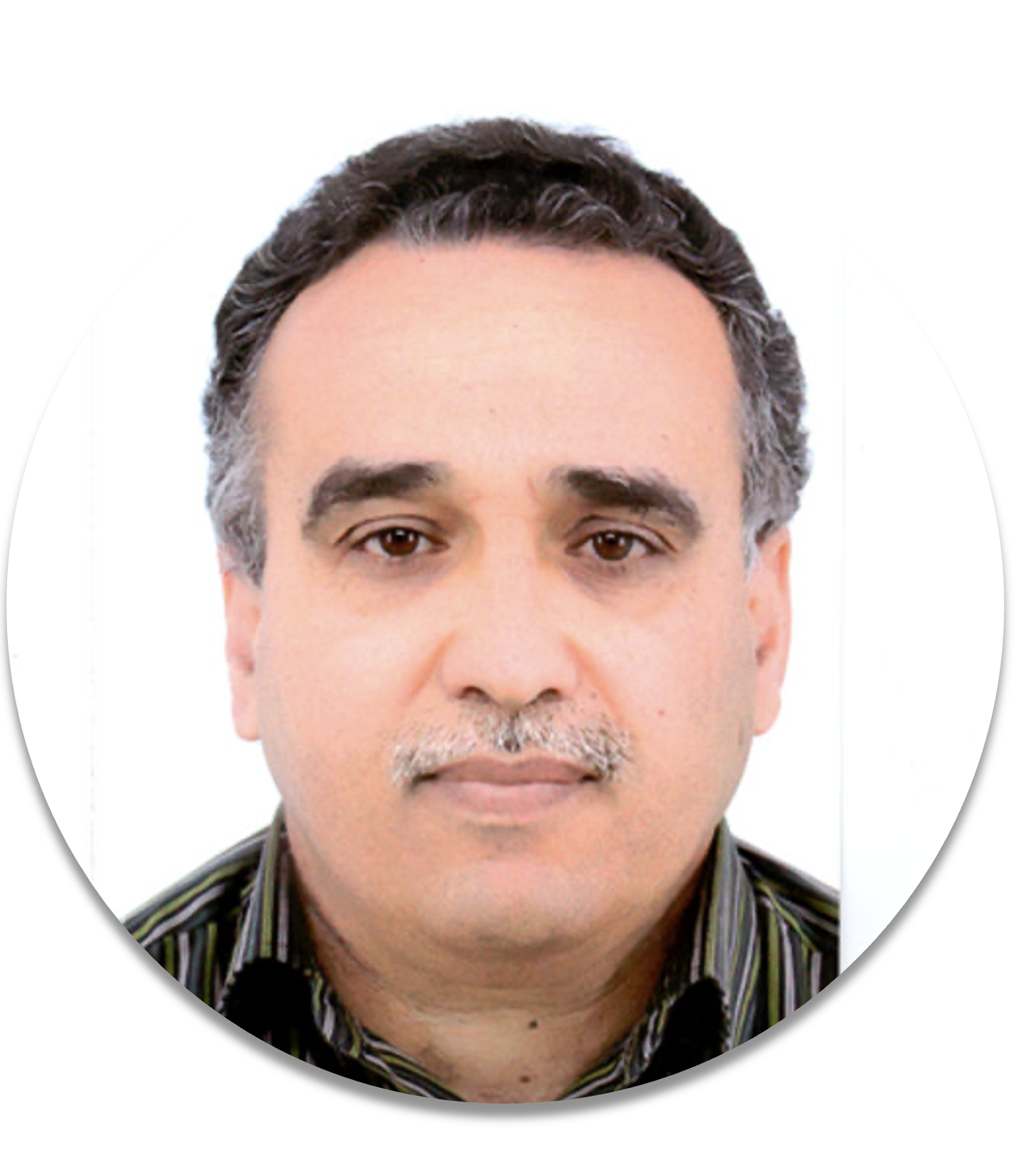}\inst{1}  \\and 
\AuthorWithPic[6mm][0.18em]{Islem Rekik}{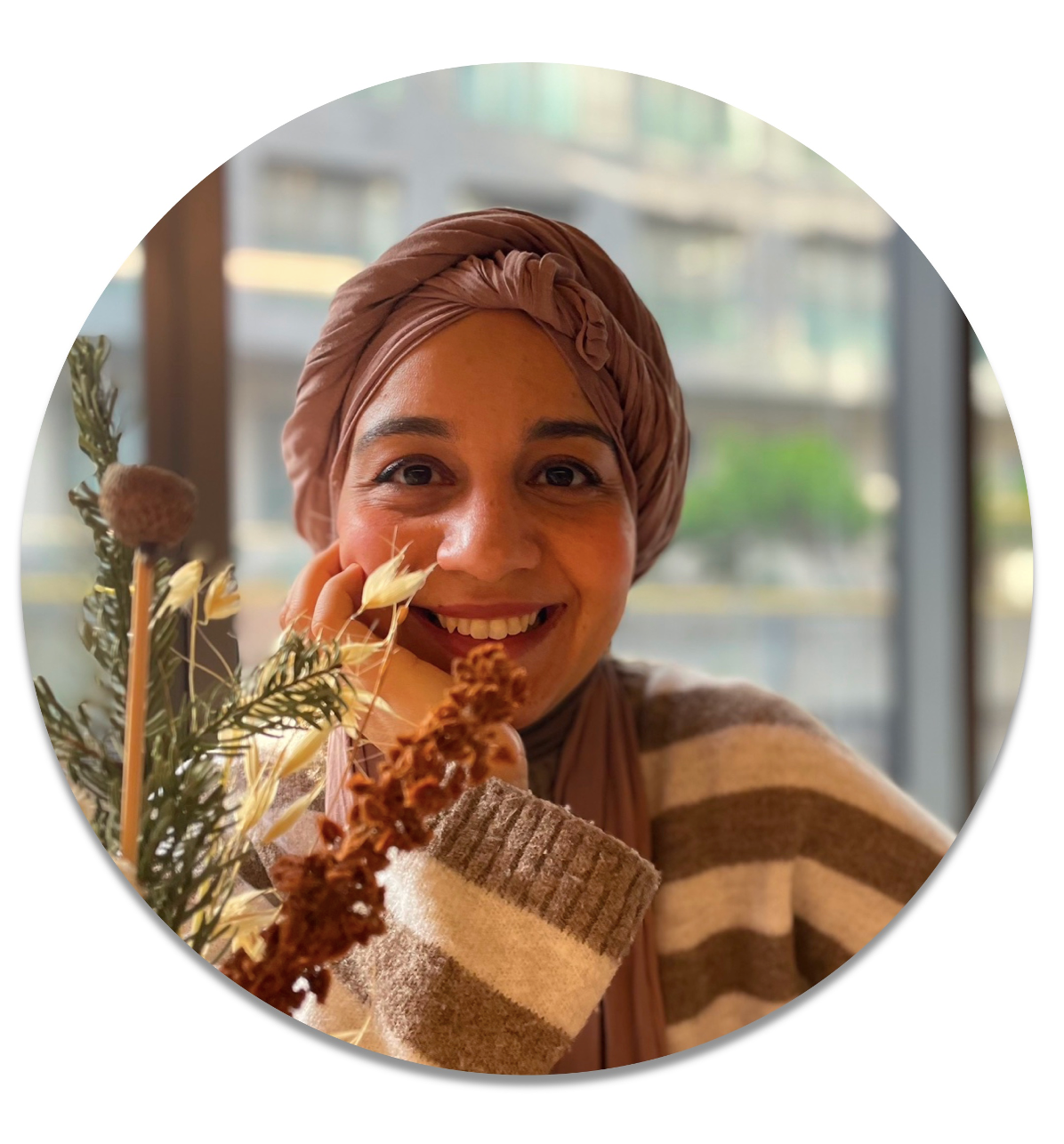}\inst{2}\orcidA{} \thanks{ {corresponding author: i.rekik@imperial.ac.uk, 
\url{http://basira-lab.com}. }}}

\authorrunning{Soussia et al.}
%
\institute{National Engineering School of Sousse, University of Sousse, LATIS- Laboratory of Advanced Technology and Intelligent Systems, 4023, Sousse, Tunisia\and
BASIRA Lab, Imperial-X and Department of Computing, Imperial College London, UK  \\
 }
\maketitle              
\begin{abstract}
The generation of connectional brain templates (CBTs) has recently garnered significant attention for its potential to identify unique connectivity patterns shared across individuals. However, existing methods for CBT learning such as conventional machine learning and graph neural networks (GNNs) are hindered by several limitations. These include: (i) poor interpretability due to their black-box nature, (ii) high computational cost, and iii) an exclusive focus on structure and topology, overlooking the cognitive capacity of the generated CBT. To address these challenges, we introduce \emph{mCOCO (multi-sensory COgnitive COmputing)}, a novel framework that leverages Reservoir Computing (RC) to \emph{learn} population-level functional CBT from BOLD (Blood-Oxygen-level-Dependent) signals. RC's dynamic system properties allow for tracking state changes over time, enhancing interpretability and enabling the modeling of brain-like dynamics, as demonstrated in prior literature. By integrating multi-sensory inputs (e.g., text, audio, and visual data), \emph{mCOCO} captures not only structure and topology but also how brain regions process information and adapt to cognitive tasks such as sensory processing, all in a computationally efficient manner. Our \emph{mCOCO} framework consists of two phases: (1) mapping BOLD signals into the reservoir to derive individual functional connectomes, which are then aggregated into a group-level CBT—an approach, to the best of our knowledge, not previously explored in functional connectivity studies —and (2) incorporating \emph{multi-sensory} inputs through a cognitive reservoir, endowing the CBT with  cognitive traits. Extensive evaluations show that our mCOCO-based template significantly outperforms GNN-based CBT in terms of centeredness, discriminativeness, topological soundness, and multi-sensory memory retention. Our source code is available at \url{https://github.com/basiralab/mCOCO}.\footnote{This paper has been for publication at the main MICCAI 2025 conference. \shadedlink{https://youtu.be/tV_RFQlTpQ0}{[mCOCO YouTube Video]}.}

\keywords{Reservoir Computing  \and Connectional Brain Template \and Cognitive capacity}
\end{abstract}
\section{Introduction}

Neuroscience has traditionally focused on investigating individual brain connectivity on a  functional \cite{turk2019functional}, structural \cite{bu2021structural}, and morphological \cite{soussia2017high} levels. However, recent advancements have shifted towards learning population-level connectivity by integrating multi-graph brain networks into a unified representation: the connectional brain template (CBT) \cite{ChaariAkdagRekik2022MultigraphIntegration}. This normalized framework allows for group-level comparisons, biomarker discovery, and the tracking of evolving discriminative connectivities over time through the longitudinal CBT \cite{demirbilek2023predicting}, among many other applications \cite{chaari2021estimation,ozgur2022population}. CBT learning has traditionally relied on conventional machine learning methods, such as clustering \cite{dhifallah2020estimation} or multi-kernel manifold learning \cite{cinar2022deep}, and recent methods like graph neural networks (GNNs), including the deep graph normalizer (DGN) \cite{Gurbuz:2020}, which uses edge-conditioned GNNs to integrate multi-graph morphological networks into a single CBT. While these methods show promise, they have key limitations. They are often considered "black boxes" \cite{rudin2019stop}, hindering interpretability of the learned CBT. Additionally, they are computationally intensive, demanding significant processing power and memory, especially with large datasets. Finally, they focus primarily on the structure and topology of the generated CBT, neglecting the \emph{cognitive} traits, such as visual and auditory processing, that are crucial to understanding human cognition.
\par
This raises the question: \emph{How can we generate a more interpretable and cognitively enhanced CBT with minimal computational cost?} Addressing such question is of paramount in developing more holistic templates that not only capture the topology and structure of networks but also model how brain regions process cognitive information. Such advancements would provide deeper insights and stronger comparisons between healthy and disordered populations. To address this gap, we propose a novel \emph{multi-sensory COgnitive COmputing (mCOCO)} framework to learn population-level CBT from BOLD signals endowed with cognitive traits. Specifically, we adopt RC \cite{tanaka2019recent} as our foundational framework, as prior research highlights its effectiveness in mimicking brain-like behavior. For example, Enel et al. \cite{enel2016reservoir} demonstrated that the prefrontal cortex exhibits properties similar to RC. Damicelli et al. \cite{DamicelliHilgetagGoulas2022} explored the potential influence of connectivity on the performance of artificial neural networks across different primate species using RC. Additionally, Xiao et al. \cite{xiao2024dyngnn} integrated RCs trained on dynamic memory tasks into GNNs for 4D brain connectivity forecasting. Nevertheless, these methods primarily focused on random sequences and did not incorporate sensory inputs such as vision or audio. One notable study incorporating sensory processing is Katori et al. \cite{katori2024brain}, which processed time-varying sensory signals and integrated feedback for adaptive decision-making using predictive coding and reinforcement learning. 

\begin{figure}
    \centering
    \includegraphics[width=1\linewidth]{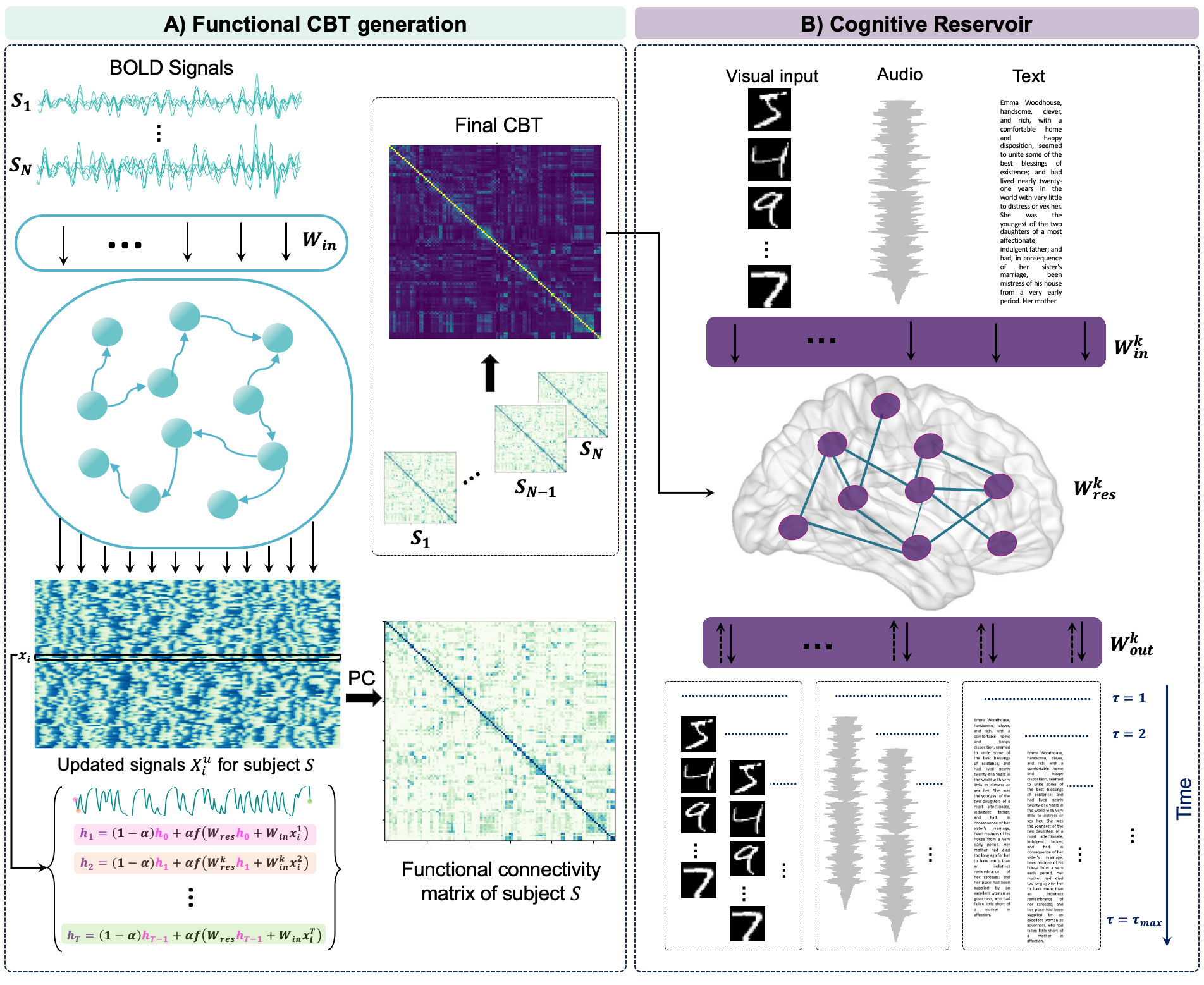}
    \caption{\textit{Overview of the mCOCO pipeline for generating cognitively enhanced CBTs.} \textbf{A) Functional CBT Generation:} BOLD signals are processed through a fixed random reservoir, producing reservoir-encoded signals. Pearson correlation is then applied to construct subject-level functional connectomes, which are aggregated into a population-level CBT. \textbf{B) Cognitive Reservoir:} The resulting CBT is instantiated in a cognitive reservoir that receives multi-sensory inputs (visual, auditory, textual), to assess its memory capacity by predicting delayed input signals.}

    \label{fig:main}
\end{figure}

Our approach goes beyond sensory processing. To the best of our knowledge, we are the first to use RC to learn individual and population-level functional connectomes from BOLD signals. By projecting temporal signals into a high-dimensional state space, the reservoir efficiently captures non-linear relationships \cite{schrauwen2007overview} through its recurrent dynamics. \textbf{We hypothesize that our \emph{mCOCO} framework will yield a more interpretable and cognitively enhanced CBT}.\label{hypothesis} Unlike fully connected neural networks, where all weights are optimized, RC only trains the output weights, making its learning process more transparent and interpretable. We present three major contributions. On a \emph{methodological level}, our \emph{mCOCO} framework is the first RC-based approach to generate cognitively enhanced brain templates. It consists of two phases (Fig.\ref{fig:main}): \textbf{1)} we use a random reservoir to map BOLD signals into a higher-dimensional space to derive individual functional connectomes, which are then aggregated to obtain the group-level connectome, \textbf{2)} We incorporate multi-sensory inputs into a cognitive reservoir to investigate memory retention across different modalities. On a \emph{conceptual level,} our framework introduces the concept of \emph{cognitive CBT} generation from a population of subjects, each represented by their BOLD signals. On a \emph{clinical level}, we demonstrate that \emph{mCOCO} generates reliable and biologically sound brain templates that capture the cognitive traits of autistic subjects and reveal key differences when compared to typically developing subjects across different sensory inputs.

\section{Method}

 In this section, we present the \emph{mCOCO} framework, illustrated in Fig.\ref{fig:main}, consisting of two components: (1) functional CBT generation from BOLD signals using a random reservoir  (Fig.\ref{fig:main}.A) and (2) a cognitive reservoir fed with multi-sensory input for cognitively enhanced CBTs (Fig.~\ref{fig:main}.B). 

\subsubsection{Functional CBT generation.}
The first block in our framework (Fig.\ref{fig:main}.A) encodes the process of generating functional CBT using a random reservoir. Given a set of N training subjects, each subject $S$ is represented by BOLD signals $\mathbf{X}_i \in \mathbb{R}^{R \times T}$ where R is the number of regions of interest (ROIs) and T is the number of timepoints. These signals represent the time-varying neural activity across different ROIs in the brain. For each subject, the signals are mapped onto a high dimensional reservoir space through the input weight matrix $W_{in}$. By feeding them into the reservoir, the network's recurrent dynamics capture non-linear temporal dependencies between brain regional activities that might not be immediately depicted in the raw signals. The internal reservoir states $\mathbf{h}(t)$ are updated iteratively as follows: 
\begin{equation}
\mathbf{h}(t+1) = (1 - \alpha
_r) \mathbf{h}(t) + \alpha
_r\tanh\left( \mathbf{W}_{\text{in}} \mathbf{X}_i(t+1) + \mathbf{W}_{\text{res}} \mathbf{h}(t) \right)
\end{equation}
where $\mathbf{h}(t) \in \mathbb{R}^{M}$ is the reservoir state vector at time $t$, with $M$ being the number of neurons in the reservoir. $\mathbf{W}_{\text{in}} \in \mathbb{R}^{M \times R}$ is the input weight matrix that projects the BOLD signals into the reservoir, $\mathbf{W}_{\text{res}} \in \mathbb{R}^{M \times M}$ is the recurrent weight matrix that governs the internal dynamics of the reservoir, $\alpha_r \in [0, 1]$ is the leak rate, which controls the balance between retaining previous states and incorporating new inputs.
Once the reservoir states are stabilized over all T time points, we obtain the learned subject signals \( \mathbf{X}_i^u \in \mathbb{R}^{R \times T} \), where each time point of the new signal is derived from the accumulated reservoir state vector \( \mathbf{h}(t) \) (Fig.\ref{fig:main}.A). This process captures the temporal evolution of neural activity, encoding both short-term and long-term dependencies. Next, we compute the functional connectivity for each subject (Fig.~\ref{fig:main}.A) by calculating the Pearson correlation coefficient between the final learned signal states \( \mathbf{X}_i^u \) of each pair of ROIs. By transforming raw BOLD signals into reservoir states, we generate a richer representation of brain activity that captures both temporal dynamics and non-linear interactions. This improves the robustness of functional connectivity computation, unlike conventional methods that assume linearity and ignore non-linear dynamics. We apply this process to each subject, producing an individual functional connectivity matrix. These matrices are then mean aggregated to form a population-level CBT (Fig. \ref{fig:main}.A).

\subsubsection{Proposed Multi-Sensory Cognitive Reservoir.}

In the second stage of our framework (Fig.~\ref{fig:main}.B), we incorporate multi-sensory inputs \( \{\mathbf{P}_k\}_{k=1}^c \), where \(c\) is the number of modalities. We adopt an Echo State Network (ESN) architecture \cite{jaeger2001echo}, using the generated functional CBT as the reservoir. The CBT serves as a recurrent network to capture cognitive functionalities, processing diverse sensory inputs (e.g., visual, auditory, and language signals). Each modality is represented by an input \( \mathbf{P}_k \in \mathbb{R}^{D_{in,k}} \), with dimensionality \( D_{in,k} \), and fed into the reservoir via input weight matrix \( \mathbf{W}_{\text{in}} \in \mathbb{R}^{R \times D_{in,k}} \). The recurrent matrix \( \mathbf{W}_{\text{res}} \in \mathbb{R}^{R \times R} \) processes inputs over time, updating the reservoir state \( \mathbf{h}^k(t) \) based on the current input and previous state. The leak rate \( \alpha_p \) controls the balance between these components. The update rule is given by:

\begin{equation}
 \mathbf{h}^k(t) = \tanh \left( \alpha_p \mathbf{W}^k_{\text{in}} \mathbf{P}_k(t) + (1 - \alpha_p) \mathbf{W}^k_{\text{res}} \mathbf{h}^k(t-1) \right), \space \space    k \in \{1,\dots,c\}
\end{equation}

To evaluate the reservoir's ability to retain and recall information over time, we train it to predict delayed versions of its input. Specifically, for a given time lag \( \tau \), the reservoir learns to reconstruct the input from \( t-\tau \) based on the current state at time \( t \). This involves predicting \( \mathbf{P}_k(t-\tau) \) given \( \mathbf{h}^k(t) \). For example, if the input is a sequence of MNIST digits (e.g., \([3,4,5,6]\)), the reservoir learns to predict \([0,3,4,5]\) when given \([3,4,5,6]\) as input, where the first prediction is set to zero due to the absence of prior input before the sequence starts. By training the reservoir to predict these delayed inputs, we are effectively teaching it to model memory. Memory, in this context, is time-dependent, meaning the brain network has the ability to recall information over different time lags. We aim to evaluate and learn this temporal memory through the reservoir's prediction task. Finally, the predicted output for the \( k \)-th modality at time \( t \) is computed using the readout weights \( \mathbf{W}^k_{\text{out}} \in \mathbb{R}^{R \times D_{out,k}} \) as:

\begin{equation}
\hat{\mathbf{P}}_k(t) = \mathbf{W}^k_{\text{out}} \mathbf{h}^k(t)
\end{equation}

The goal is to minimize the error between the predicted output \( \hat{\mathbf{P}}_k(t) \) and delayed version of the true input \( \mathbf{P}_k(t-\tau) \). Thereby, endowing the brain template with a multi-sensory recall capacity. The error can be quantified using a loss function, typically the mean squared error (MSE) across \( T_p \) time steps:

\begin{equation}
\mathcal{L} = \frac{1}{T_p} \sum_{t=1}^{T_p} \left( \hat{\mathbf{P}}_k(t) - \mathbf{P}_k(t-\tau) \right)^2
\end{equation}

\section{Experiments and Results}
\textbf{Dataset.} In this study, we used a dataset of BOLD signals downloaded from the Autism Brain Imaging Data Exchange (ABIDE) website \footnote{\url{https://preprocessed-connectomes-project.org/abide/download.html}}. The dataset includes a total of 884 subjects from all sites, comprising 408 individuals diagnosed with Autism Spectrum Disorder (ASD) and 476 typically developing (TD) controls. Preprocessing of the ABIDE data was performed using version X of the Configurable Pipeline for the Analysis of Connectomes (C-PAC). The BOLD signals were parcellated using the Harvard-Oxford Atlas (HOA) into 111 ROIs based on anatomical structures.

\noindent\textbf{Hyperparameter tuning}. For the CBT generation, we used a random reservoir with \( M = 111 \) neurons. The leak rate was set to \( \alpha_r = 0.5 \), and the spectral radius was fixed at \( 1.45 \) to ensure stable dynamics. The reservoir weight matrix \( \mathbf{W}_{\text{res}} \) and the input weight matrix \( \mathbf{W}_{\text{in}} \) were sampled from a uniform distribution in \([-1, 1]\). To evaluate the generalizability of our model, we performed 5-fold cross-validation.

\noindent\textbf{Sensory input}. We used three cognitive inputs: \textit{i) the visual input}, we selected 100 samples from the MNIST dataset \cite{deng2012mnist}, downsampled to \( 15 \times 15 \) pixels, normalized to \([-1, 1]\), and flattened into 225-dimensional vectors. \textit{ii) the auditory input}, we used segments from Beethoven’s \emph{Ode to Joy} and a \emph{Quranic recitation}, both transformed into Mel-Frequency Cepstral Coefficients (MFCCs) via the \texttt{librosa} library. \textit{iii) the textual input}, we used the Gutenberg corpus from the \texttt{NLTK} library, generating word embeddings with a trained Word2Vec model. For all modalities, the input data was split into 80-20 training and testing sets, with a maximum time lag \( \tau_{\text{max}} = 20 \). The ESN was configured with spectral radius \( \rho = 0.99 \), input scaling \( \epsilon = 1 \), and leak rate \( \alpha_p = 1 \), and implemented using the \texttt{echoes} library \cite{damicelli2019echoes}.

\noindent\textbf{Benchmark method.} We selected DGN \cite{Gurbuz:2020} as our benchmark method, as it outperformed seven integration models for generating CBTs, as shown in the comparative survey \cite{ChaariAkdagRekik2022MultigraphIntegration}. This positions DGN as a state-of-the-art method for our evaluation.

\subsection{CBT evaluation} In this work, we generated two distinct CBTs (ASD and TD) and evaluated their centeredness, cognitive capacity, discriminativeness and topological soundness.

\begin{figure}
    \centering
    \includegraphics[width=0.8\linewidth]{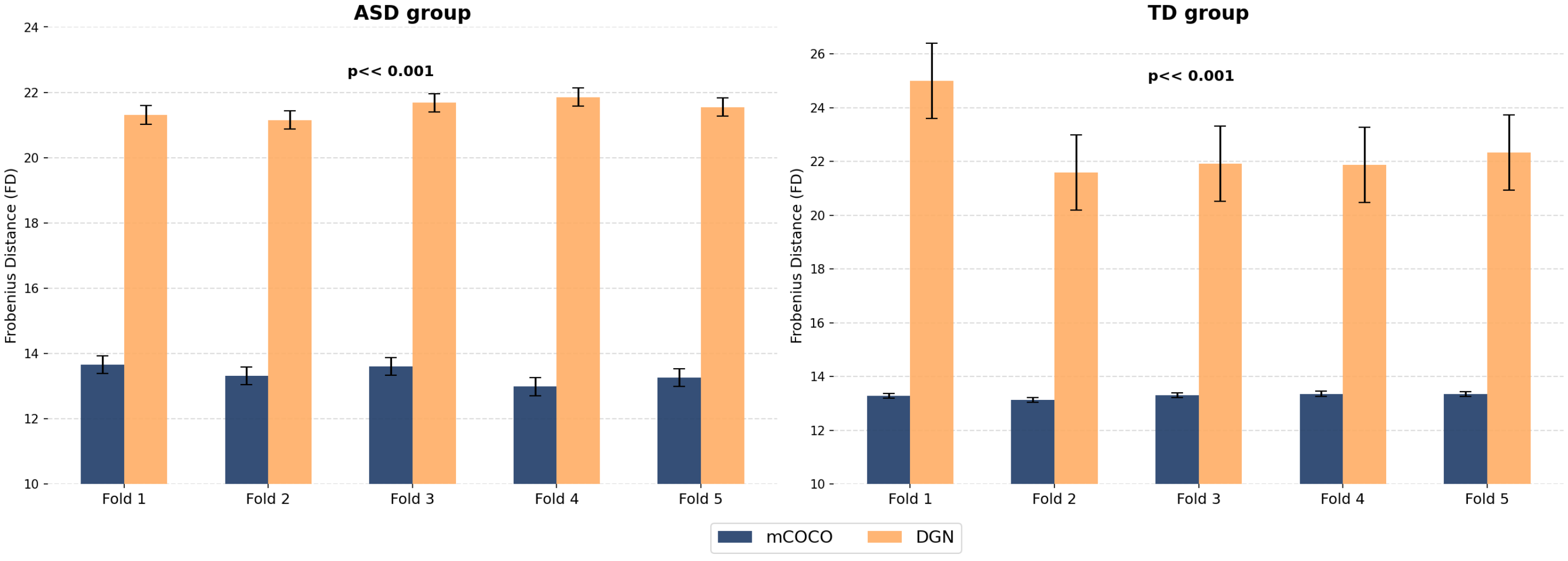}
    \caption{\textit{Centeredness comparison between CBTs using mean Frobenius distance.}}
    \label{fig:centeredness}
\end{figure}

\subsubsection{Centeredness.} To assess the CBT centeredness, we quantify the deviation of the learned templates from the left-out subjects using the mean Frobenius distance $d_F(\mathbf{C}, \mathbf{M}) = \frac{1}{N_{test}} \sum_{j=1}^{N_{test}} \| \mathbf{C} - \mathbf{M}_j \|_F$. This measures the discrepancy between the generated CBT \( \mathbf{C} \) and the functional connectivity matrices \( \mathbf{M}_j \) of the testing subjects, evaluating how well the CBT generalizes to unseen data. The dataset was split into training and testing sets using 5-fold cross-validation, with the CBT trained on the training set and centeredness assessed on the left-out samples. Our results show that the \emph{mCOCO}-based CBT significantly outperforms the DGN-based CBT across all folds (Fig.~\ref{fig:centeredness}) as indicated by the p-value (p $<< 0.001$). This superior performance is due to \emph{mCOCO}'s ability to capture dynamic temporal dependencies in the BOLD signals, which are key to modeling complex connectivity patterns in ASD and TD populations.

\begin{figure}
    \centering
    \includegraphics[width=1\linewidth]{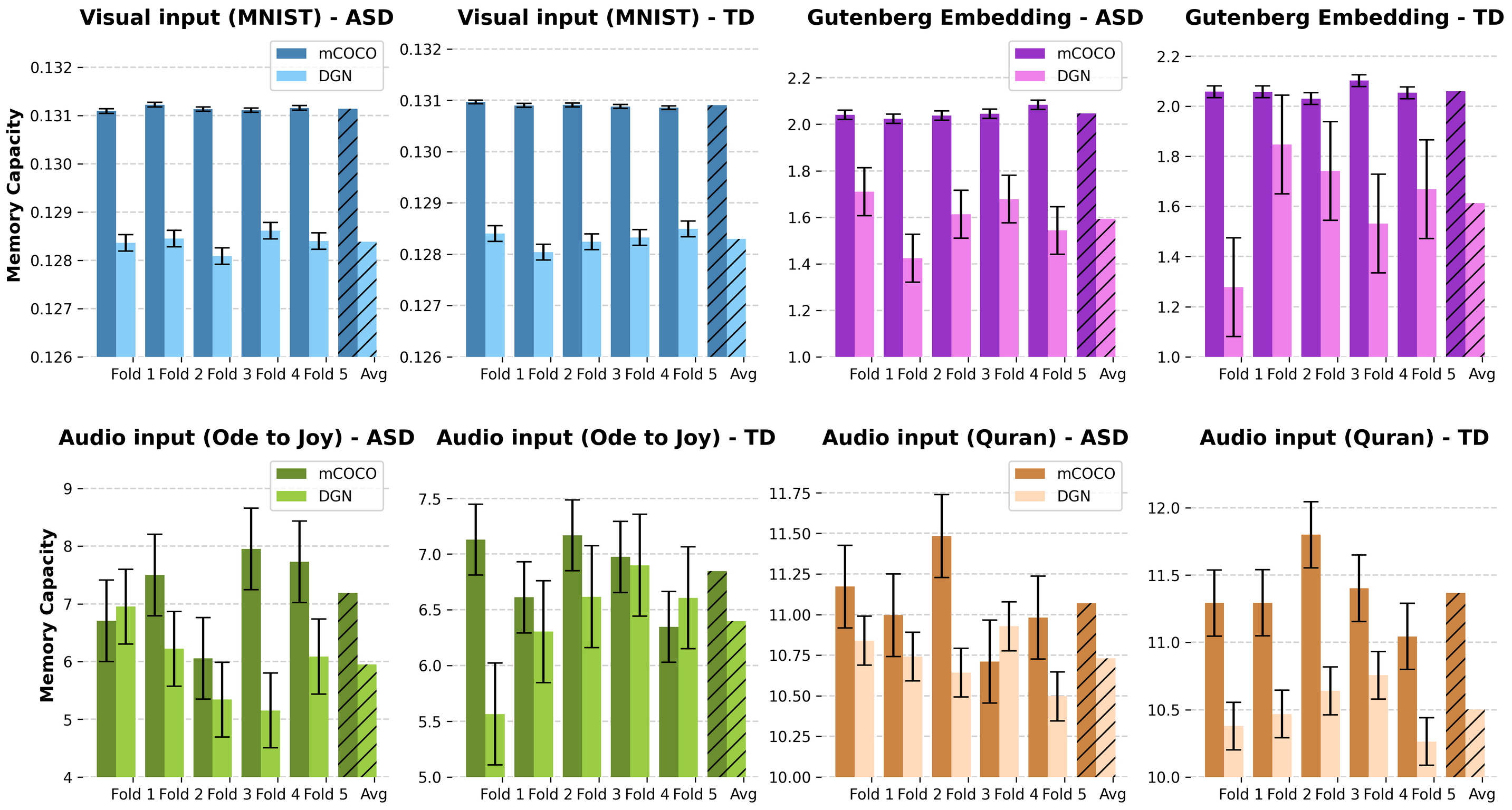}
    \caption{\textit{Comparison of memory capacity across various input types (Visual, Textual and Audio).} Results show the performance across five folds, with the final bar representing the average and error bars indicating variability.}
    \label{fig:cog}
\end{figure}

\subsubsection{Cognitive Capacity.}
To evaluate the cognitive capacity of the generated CBTs, we assessed their ability to retain and recall information across various sensory modalities: visual, textual, and auditory. This was achieved by measuring the \textit{memory capacity (MC)}, which quantifies the reservoir's ability to reconstruct delayed versions of its input at different time lags. MC is computed as the cumulative score of squared Pearson correlation coefficients ($\rho$) between the true delayed input $\mathbf{P}_{k,\tau}$ and the predicted output $\mathbf{\hat{P}}_{k}$, defined as $\text{MC} = \sum_{\tau=1}^{\tau_{max}}\rho^2(\mathbf{P}_{k,\tau}, \mathbf{\hat{P}}_{k})$. Higher MC values indicate better recall, reflecting superior cognitive capacity. The \emph{mCOCO}-based CBT consistently outperforms the DGN-based CBT, particularly for text and audio, supporting our hypothesis that \emph{mCOCO} generates more cognitive templates. \textit{The visual stimuli}, represented by handwritten digits from the MNIST dataset, show the lowest MC values ($<1$) across both ASD and TD groups. This is likely because the lack of temporal complexity does not fully engage the reservoir's nonlinear dynamics. Visual processing primarily involves the hierarchical and parallel processing of features in areas like the primary visual cortex and higher-order regions, which may not align well with the sequential nature of RC. Exploring techniques like Hebbian learning and intrinsic plasticity \cite{schrauwen2007overview} could enhance RC performance in visual tasks. In contrast, \textit{the text input} demonstrates higher memory capacity ($MC > 2$) values compared to MNIST. These higher values indicate that the \emph{mCOCO}-based CBT effectively retains the semantic and syntactic structure of the input text, allowing the model to reconstruct earlier parts of a sentence across various time lags. Individuals with ASD often struggle to integrate linguistic contexts, which may reduce their retention of sequential information during text processing \cite{williams2013brain}. This aligns with the lower average MC values observed in ASD compared to TD. Finally, \textit{the audio input} achieves the highest MC values ($MC > 7$ for \textit{Ode to Joy} and $MC > 11$ for \textit{Quranic recitation}), consistent with the temporal richness of auditory signals, which engage the reservoir's nonlinear dynamics. The higher MC for Quran audio may stem from the linguistic and prosodic rules that make the content more predictable and easier for the reservoir to model. This suggests that spiritually or culturally significant auditory stimuli, like religious texts or prayers, may be processed and recalled more easily\cite{dolcos2004interaction,bower1978emotional}. The increased MC values may also reflect the emotional salience of spiritual content, which enhances attention and memory encoding, as the brain prioritizes emotionally impactful information \cite{dolcos2004interaction}. Spiritual stimuli often involve deeper cognitive processing and stronger connections to personal beliefs, leading to better retention and recall \cite{vaghefi2015spirituality}.

\subsubsection{CBT discriminativeness.}To evaluate the discriminative power of generated CBTs, we used a CBT-shot learning approach with an SVM classifier. The SVM was trained on CBTs from two classes (e.g., ASD and TD) and evaluated on left-out subjects using 5-fold cross-validation.The results in Table~\ref{tab:discriminative_performance} show that the SVM trained with \emph{mCOCO}-CBTs outperforms the one trained with DGN-CBTs, achieving 62.02\% accuracy, significantly higher than the 51.3\% achieved with DGN-CBTs.

\begin{table}[h!]
\centering
\label{tab:discriminative_performance}
\caption{Performance of SVM trained on CBTs generated by mCOCO and DGN.}
\setlength{\tabcolsep}{6pt}
\begin{tabular}{ccccc}
\hline
\textbf{Model} & \textbf{Accuracy} & \textbf{Sensitivity} & \textbf{Specificity} & \textbf{F1} \\ \hline
SVM-DGN \cite{Gurbuz:2020} & 51.36\% & \textbf{67.27}\% & 32.93\% & 0.27 \\ 
SVM-mCOCO & \textbf{62.02}\% & 42.93\% & \textbf{81.11}\% & \textbf{0.52} \\ \hline

\end{tabular}

\end{table}

\begin{figure}
    \centering
    \includegraphics[width=1\linewidth]{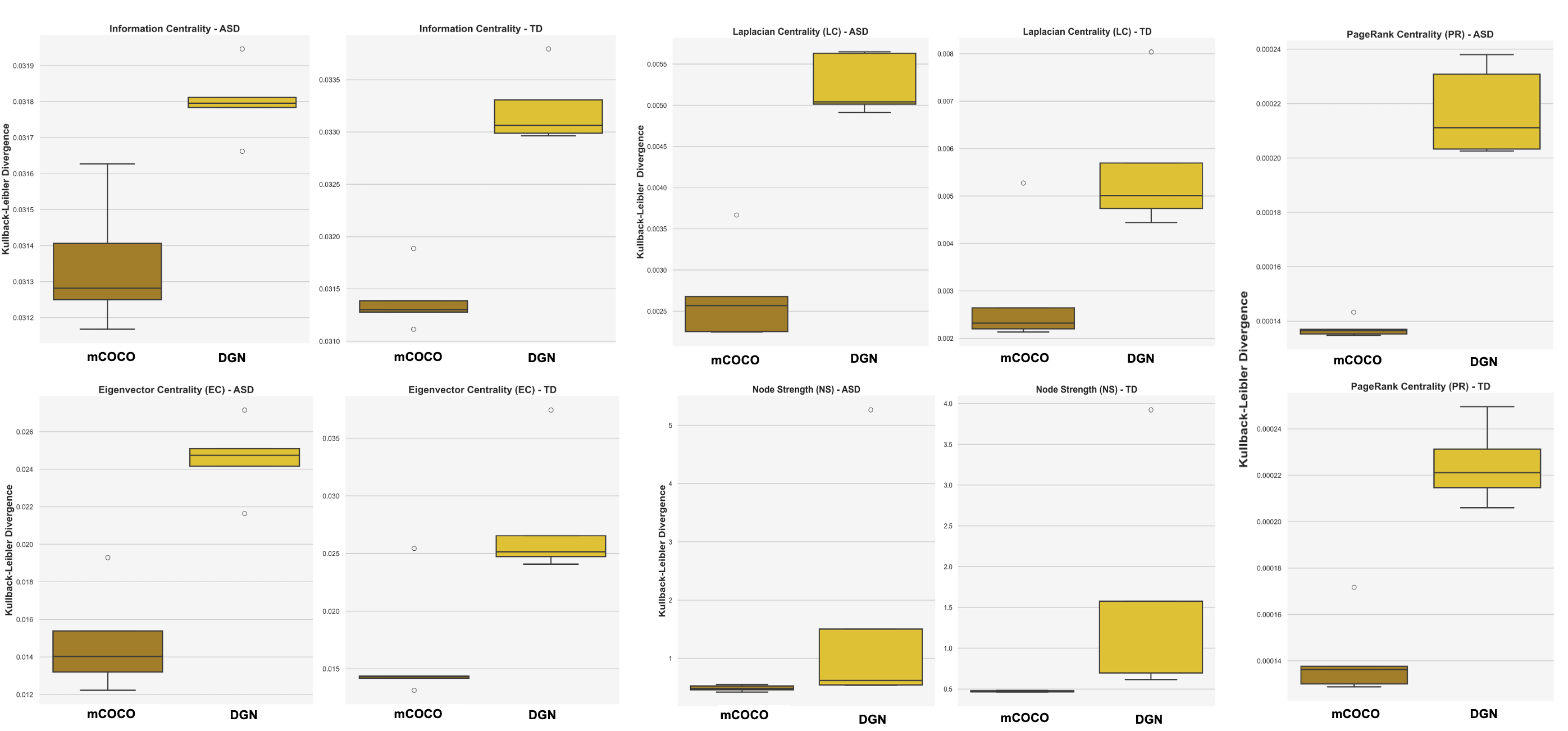}
    \caption{\textit{Average of Kullback-Liebler divergence distribution across 5-fold cross validation between a testing set and generated CBTs by mCOCO and DGN\cite{Gurbuz:2020} respectively.}}
    \label{fig:topology}
\end{figure}

\subsubsection{Topological Soundness.} Here, we examined the distribution discrepancies of key topological measures (information centrality, Laplacian centrality, eigenvector centrality, PageRank centrality, and node strength) between the generated CBTs and a set of left-out subjects. To quantify these discrepancies, we utilized the Kullback-Leibler (KL) divergence which measures the dissimilarity between two probability distributions. The results, displayed as box plots in Fig.~\ref{fig:topology}, show that our \emph{mCOCO}-based CBT consistently exhibits lower divergence across all evaluated measures compared to the DGN-based template. This finding suggests that the \emph{mCOCO}-based CBT more accurately reflects topological properties of the brain networks.

\section{Conclusion}

We introduced a novel multi-sensory COgnitive COmputing (mCOCO) framework, generating a functional population-driven CBT from BOLD signals using random RC while enhancing its cognitive capacity with multi-sensory inputs. Our framework produced a well-centered, topologically sound, and cognitively enhanced CBT, setting a new benchmark in cognitive generative modeling. While this work focused on pairwise brain region interactions, future work will extend it to capture higher-order interactions, improving the representation of complex brain networks and cognitive functions.

%

\bibliographystyle{splncs}
\bibliography{Biblio}

\begin{thebibliography}{10}

\bibitem{soussia2017high}
Soussia, M., Rekik, I.:
\newblock High-order connectomic manifold learning for autistic brain state identification.
\newblock In: Connectomics in NeuroImaging: First International Workshop, CNI 2017, Held in Conjunction with MICCAI 2017, Quebec City, QC, Canada, September 14, 2017, Proceedings 1, Springer (2017)  51--59

\bibitem{turk2019functional}
Turk, E., Van Den~Heuvel, M.I., Benders, M.J., De~Heus, R., Franx, A., Manning, J.H., Hect, J.L., Hernandez-Andrade, E., Hassan, S.S., Romero, R.,  et~al.:
\newblock Functional connectome of the fetal brain.
\newblock Journal of Neuroscience \textbf{39} (2019)  9716--9724

\bibitem{bu2021structural}
Bu, X., Cao, M., Huang, X., He, Y.:
\newblock The structural connectome in adhd.
\newblock Psychoradiology \textbf{1} (2021)  257--271

\bibitem{ChaariAkdagRekik2022MultigraphIntegration}
Chaari, N., Camgoz~Akdag, H., Rekik, I.:
\newblock Comparative survey of multigraph integration methods for holistic brain connectivity mapping (2022) Preprint submitted to arXiv.

\bibitem{demirbilek2023predicting}
Demirbilek, O., Rekik, I., Initiative, A.D.N.,  et~al.:
\newblock Predicting the evolution trajectory of population-driven connectional brain templates using recurrent multigraph neural networks.
\newblock Medical Image Analysis \textbf{83} (2023)  102649

\bibitem{chaari2021estimation}
Chaari, N., Akda{\u{g}}, H.C., Rekik, I.:
\newblock Estimation of gender-specific connectional brain templates using joint multi-view cortical morphological network integration.
\newblock Brain Imaging and Behavior \textbf{15} (2021)  2081--2100

\bibitem{dhifallah2020estimation}
Dhifallah, S., Rekik, I., Initiative, A.D.N.,  et~al.:
\newblock Estimation of connectional brain templates using selective multi-view network normalization.
\newblock Medical image analysis \textbf{59} (2020)  101567

\bibitem{ozgur2022population}
{\"O}zg{\"u}r, O., Rekik, A., Rekik, I.:
\newblock Population template-based brain graph augmentation for improving one-shot learning classification.
\newblock arXiv preprint arXiv:2212.07790 (2022)

\bibitem{Gurbuz:2020}
Gurbuz, M.B., Rekik, I.:
\newblock Deep graph normalizer: A geometric deep learning approach for estimating connectional brain templates.
\newblock Medical Image Computing and Computer Assisted Intervention (2020)  155--165

\bibitem{rudin2019stop}
Rudin, C.:
\newblock Stop explaining black box machine learning models for high stakes decisions and use interpretable models instead.
\newblock Nature machine intelligence \textbf{1} (2019)  206--215

\bibitem{enel2016reservoir}
Enel, P., Procyk, E., Quilodran, R., Dominey, P.F.:
\newblock Reservoir computing properties of neural dynamics in prefrontal cortex.
\newblock PLoS computational biology \textbf{12} (2016)  e1004967

\bibitem{cinar2022deep}
Cinar, E., Haseki, S.E., Bessadok, A., Rekik, I.:
\newblock Deep cross-modality and resolution graph integration for universal brain connectivity mapping and augmentation.
\newblock In: MICCAI Workshop on Imaging Systems for GI Endoscopy, Springer (2022)  89--98

\bibitem{noor2014potential}
Noor, A.K.:
\newblock Potential of cognitive computing and cognitive systems.
\newblock Open Engineering \textbf{5} (2014)

\bibitem{katori2024brain}
Katori, Y.:
\newblock Brain-inspired reservoir computing models.
\newblock Photonic Neural Networks with Spatiotemporal Dynamics (2024)  259

\bibitem{damicelli2019echoes}
Damicelli, F.:
\newblock echoes: Echo state networks with python.
\newblock \url{https://github.com/fabridamicelli/echoes} (2019)

\bibitem{tanaka2019recent}
Tanaka, G., Yamane, T., H{\'e}roux, J.B., Nakane, R., Kanazawa, N., Takeda, S., Numata, H., Nakano, D., Hirose, A.:
\newblock Recent advances in physical reservoir computing: A review.
\newblock Neural Networks \textbf{115} (2019)  100--123

\bibitem{schrauwen2007overview}
Schrauwen, B., Verstraeten, D., Van~Campenhout, J.:
\newblock An overview of reservoir computing: theory, applications and implementations.
\newblock In: Proceedings of the 15th european symposium on artificial neural networks. p. 471-482 2007. (2007)  471--482

\bibitem{DamicelliHilgetagGoulas2022}
Damicelli, F., Hilgetag, C.C., Goulas, A.:
\newblock Brain connectivity meets reservoir computing.
\newblock PLOS Computational Biology (2022)

\bibitem{suarez2024connectome}
Su{\'a}rez, L.E., Mihalik, A., Milisav, F., Marshall, K., Li, M., V{\'e}rtes, P.E., Lajoie, G., Misic, B.:
\newblock Connectome-based reservoir computing with the conn2res toolbox.
\newblock Nature Communications \textbf{15} (2024)  656

\bibitem{xiao2024dyngnn}
Xiao, S., Rekik, I.:
\newblock Dyngnn: Dynamic memory-enhanced generative gnns for predicting temporal brain connectivity.
\newblock In: International Workshop on PRedictive Intelligence In MEdicine, Springer (2024)  111--123

\bibitem{jaeger2001echo}
Jaeger, H.:
\newblock The “echo state” approach to analysing and training recurrent neural networks-with an erratum note.
\newblock Bonn, Germany: German National Research Center for Information Technology GMD Technical Report \textbf{148} (2001) ~13

\bibitem{deng2012mnist}
Deng, L.:
\newblock The mnist database of handwritten digit images for machine learning research [best of the web].
\newblock IEEE signal processing magazine \textbf{29} (2012)  141--142

\bibitem{menon2013developmental}
Menon, V.:
\newblock Developmental pathways to functional brain networks: emerging principles.
\newblock Trends in cognitive sciences \textbf{17} (2013)  627--640

\bibitem{tovee2008introduction}
Tov{\'e}e, M.J.:
\newblock An introduction to the visual system.
\newblock Cambridge University Press (2008)

\bibitem{friederici2011brain}
Friederici, A.D.:
\newblock The brain basis of language processing: from structure to function.
\newblock Physiological reviews \textbf{91} (2011)  1357--1392

\bibitem{williams2013brain}
Williams, D.L., Cherkassky, V.L., Mason, R.A., Keller, T.A., Minshew, N.J., Just, M.A.:
\newblock Brain function differences in language processing in children and adults with autism.
\newblock Autism Research \textbf{6} (2013)  288--302

\bibitem{vaghefi2015spirituality}
Vaghefi, M., Nasrabadi, A.M., Golpayegani, S.M.R.H., Mohammadi, M.R., Gharibzadeh, S.:
\newblock Spirituality and brain waves.
\newblock Journal of medical engineering \& technology \textbf{39} (2015)  153--158

\bibitem{kojovic2019sensory}
Kojovic, N., Ben~Hadid, L., Franchini, M., Schaer, M.:
\newblock Sensory processing issues and their association with social difficulties in children with autism spectrum disorders.
\newblock Journal of clinical medicine \textbf{8} (2019)  1508

\bibitem{dolcos2004interaction}
Dolcos, F., LaBar, K.S., Cabeza, R.:
\newblock Interaction between the amygdala and the medial temporal lobe memory system predicts better memory for emotional events.
\newblock Neuron \textbf{42} (2004)  855--863

\bibitem{bower1978emotional}
Bower, G.H., Monteiro, K.P., Gilligan, S.G.:
\newblock Emotional mood as a context for learning and recall.
\newblock Journal of verbal learning and verbal behavior \textbf{17} (1978)  573--585

\end{thebibliography}
\end{document}